%% file: main.tex
\documentclass[sigconf]{acmart}

\usepackage{booktabs} 
\usepackage{amsmath}
\usepackage{amssymb}
\usepackage{graphicx}
\usepackage[colorinlistoftodos]{todonotes}
\usepackage{tabularx}
\usepackage{soul}

\usepackage[T1]{fontenc}
\usepackage[utf8]{inputenc}
\usepackage{caption, color, enumerate, epsfig, float, graphics, multirow, subfigure, verbatim, xspace, wrapfig, url}
\usepackage{makecell}
\usepackage[noend]{algorithmic}
\usepackage[boxed]{algorithm}
\usepackage{hyperref}
\usepackage{float}
\usepackage{diagbox}
\usepackage{makecell}
\usepackage{gensymb}
\usepackage{enumitem}
\usepackage{multirow}
\usepackage{pifont}
\usepackage{appendix}

\renewcommand\footnotetextcopyrightpermission[1]{}

\setlist[itemize]{leftmargin=*}

\newtheorem{problem}{Problem}

\newcommand{\nop}[1]{}

\newcommand{\methodFont}{\textsf}
\newcommand{\fixedtime}{\methodFont{Fixedtime}\xspace}
\newcommand{\SOTL}{\methodFont{SOTL}\xspace}
\newcommand{\NIPS}{\methodFont{DRL}\xspace}
\newcommand{\deeplight}{\methodFont{IntelliLight}\xspace}

\newcommand{\formula}{\methodFont{Formula}\xspace}

\newcommand{\p}{\mathbf{p}}
\newcommand{\q}{\mathbf{q}}

\newcommand{\ours}{\methodFont{FRAP}\xspace}

\input{preamble}
\begin{document}
\input{title}
\input{abstract}
\maketitle
\input{introduction}
\input{related_work}
\input{problem_definition}

\input{method}

\input{experiment}

\input{conclusion}

\section*{Acknowledgements}
The work was supported in part by NSF awards \#1652525, \#1618448, and \#1639150. The views and conclusions contained in this paper are
those of the authors and should not be interpreted as representing
any funding agencies.

\input{reference}
\end{document}

%% file: preamble.tex
\def\BibTeX{{\rm B\kern-.05em{\sc i\kern-.025em b}\kern-.08emT\kern-.1667em\lower.7ex\hbox{E}\kern-.125emX}}

\copyrightyear{2019}
\acmYear{2019}
\acmConference[]{ACM conference}{}{}
\acmPrice{15.00}
\acmArticle{4}

\acmDOI{10.475/123_4}
\acmISBN{123-4567-24-567/08/06}

%% file: title.tex

\title{Learning Phase Competition for Traffic Signal Control}

\renewcommand{\shortauthors}{xxx}


\author{Guanjie Zheng$^\dagger$, Yuanhao Xiong$^\ddagger$, Xinshi Zang$^\S$, Jie Feng$^\mathparagraph$, Hua Wei$^\dagger$}
\par
\author{Huichu Zhang$^\S$, Yong Li$^\mathparagraph$, Kai Xu$^\intercal$, Zhenhui Li$^\dagger$}
\affiliation{$^\dagger$Pennsylvania State University, $^\ddagger$Zhejiang Univerisity, $^\S$Shanghai Jiao Tong Univerisity, \\$^\mathparagraph$Tsinghua Univerisity, $^\intercal$Shanghai Tianrang Intelligent Technology Co., Ltd\\
$^\dagger$\{gjz5038, hzw77, jessieli\}@ist.psu.edu, $^\ddagger$xiongyh@zju.edu.cn, $^\S$zang-xs@foxmail.com, $^\S$zhc@apex.sjtu.edu.cn, $^\mathparagraph$feng-j16@mails.tsinghua.edu.cn, $^\mathparagraph$liyong07@tsinghua.edu.cn, $^\intercal$kai.xu@tianrang-inc.com}

%% file: abstract.tex

%
\begin{abstract}
Increasingly available city data and advanced learning techniques have empowered people to improve the efficiency of our city functions. Among them, improving the urban transportation efficiency is one of the most prominent topics. Recent studies have proposed to use reinforcement learning (RL) for traffic signal control. Different from traditional transportation approaches which rely heavily on prior knowledge, RL can learn directly from the feedback. On the other side, without a careful model design, existing RL methods typically take a long time to converge and the learned models may not be able to adapt to new scenarios. For example, a model that is trained well for morning traffic may not work for the afternoon traffic because the traffic flow could be reversed, resulting in a very different state representation. 

In this paper, we propose a novel design called \ours, which is based on the intuitive principle of phase competition in traffic signal control: when two traffic signals conflict, priority should be given to one with larger traffic movement (i.e., higher demand). Through the phase competition modeling, our model achieves invariance to symmetrical cases such as flipping and rotation in traffic flow. By conducting comprehensive experiments, we demonstrate that our model finds better solutions than existing RL methods in the complicated all-phase selection problem, converges much faster during training, and achieves superior generalizability for different road structures and traffic conditions.

\end{abstract}

%% file: introduction.tex

\section{Introduction}
\label{sec:introduction}

Traffic congestion is one of the most severe urban issues today, which has resulted in tremendous economic cost and waste of people's time. Congestion is caused by many factors, such as overloaded number of vehicles and bad design of road structures. Some factors may require more sophisticated policy or long-term planning. But one direct factor that could be potentially improved by today's big data and advanced learning technology is traffic signal control. 

Nowadays, the most widely used traffic signal control systems such as SCATS~\cite{scats90, SCATS} and SCOOT~\cite{hunt1982scoot, luk1984two} are still based on manually designed traffic signal plans. These plans, however, are not adaptive enough to the dynamics of today's complex traffic flows.

Recently, reinforcement learning (RL) has emerged as a promising solution to traffic signal control in real world scenarios. Unlike previous methods which rely on manually designed plans or pre-defined traffic flow models, RL methods directly learn the policy by interacting with the environment. To this end, a typical approach is to model each intersection as an agent and the agent optimizes its reward (e.g., travel time) based on the feedback received from the environment after it takes an action (i.e., setting the traffic signals). These RL approaches vary in terms of reward design (e.g., queue length~\cite{MaDH16,VaOl16,BSSN+05}, delay~\cite{BPT14,ElAb10, ElAb12,VaOl16}), state description (e.g., number of vehicles~\cite{wei2018intellilight,PraB11}, traffic image~\cite{KWBV08, BSSN+05, VaOl16,wei2018intellilight, XXL13}), learning model (e.g., deep Q-Network~\cite{VaOl16, wei2018intellilight}, policy gradient~\cite{mousavi2017traffic}, actor-critic~\cite{aslani2017adaptive, aslani2018developing}), and action design (e.g., setting the phase~\cite{AMB11, AMB14,Saca10,SCGC08}, change to next phase~\cite{VaOl16,wei2018intellilight,PBTB+13,BPT14}). 
Existing methods have shown promising results under simple traffic signal control settings, i.e., an intersection with two signal phases, where the green light is either on horizontal direction or vertical direction. 

With more complex scenarios, learning the optimal policy becomes substantially much more difficult. Consider a standard four-approach intersection where each approach has left-turn, through and right-turn traffic. There will be 8 phases (i.e., combinations of different traffic movements) according to the traffic rules (see Section~\ref{sec:problem-definition} for details). It turns out that it is much harder for the RL algorithm to deal with the 8-phase setting than the 2-phase setting. A close examination of the problem reveals that the difficulty is mainly due to the explosion of state space. In the 2-phase setting, there are only four through lanes. Assume the vehicle capacity of a lane is $n$, and the state space size of 2-phase control problem is $2\times n^4$ correspondingly (enumerating the number of vehicles on each lane, under each phase). When all eight phases are considered, four extra left-turn lanes are added and the exploration space will increase to $8 \times n^8$. Therefore, the key challenge becomes how to reduce the problem space and explore different scenarios more efficiently, so that the RL algorithm can find the optimal solution within minimal number of trials.

\begin{figure}[t]
\centering
\includegraphics[width=0.42\textwidth]{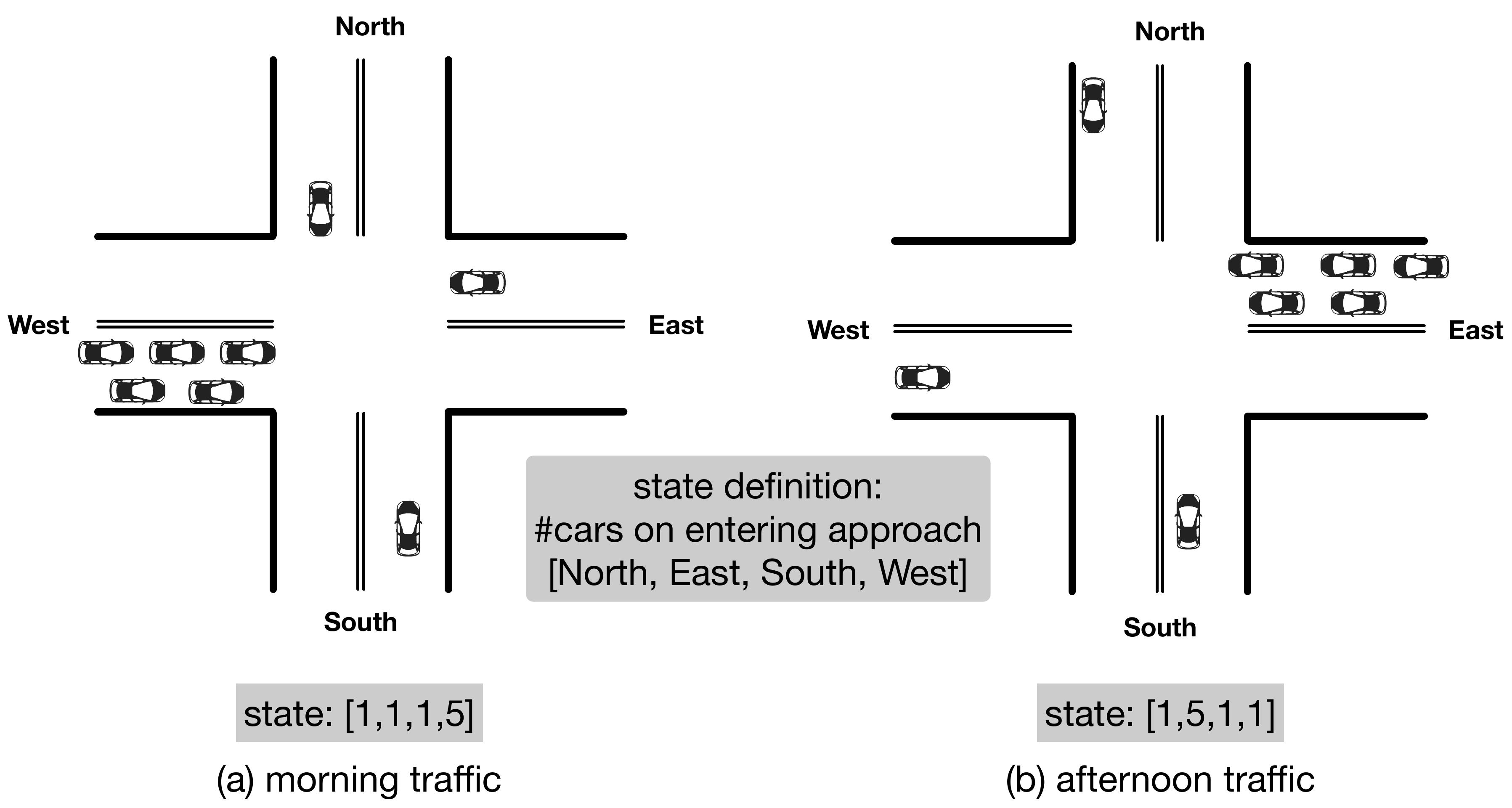}
\vspace{-2mm}
\caption{Traffic (a) and (b) are flipped cases of each other.}
\label{fig:rotate}
\vspace{-4mm}
\end{figure}

\begin{figure}[t]
\centering
\includegraphics[width=0.45\textwidth]{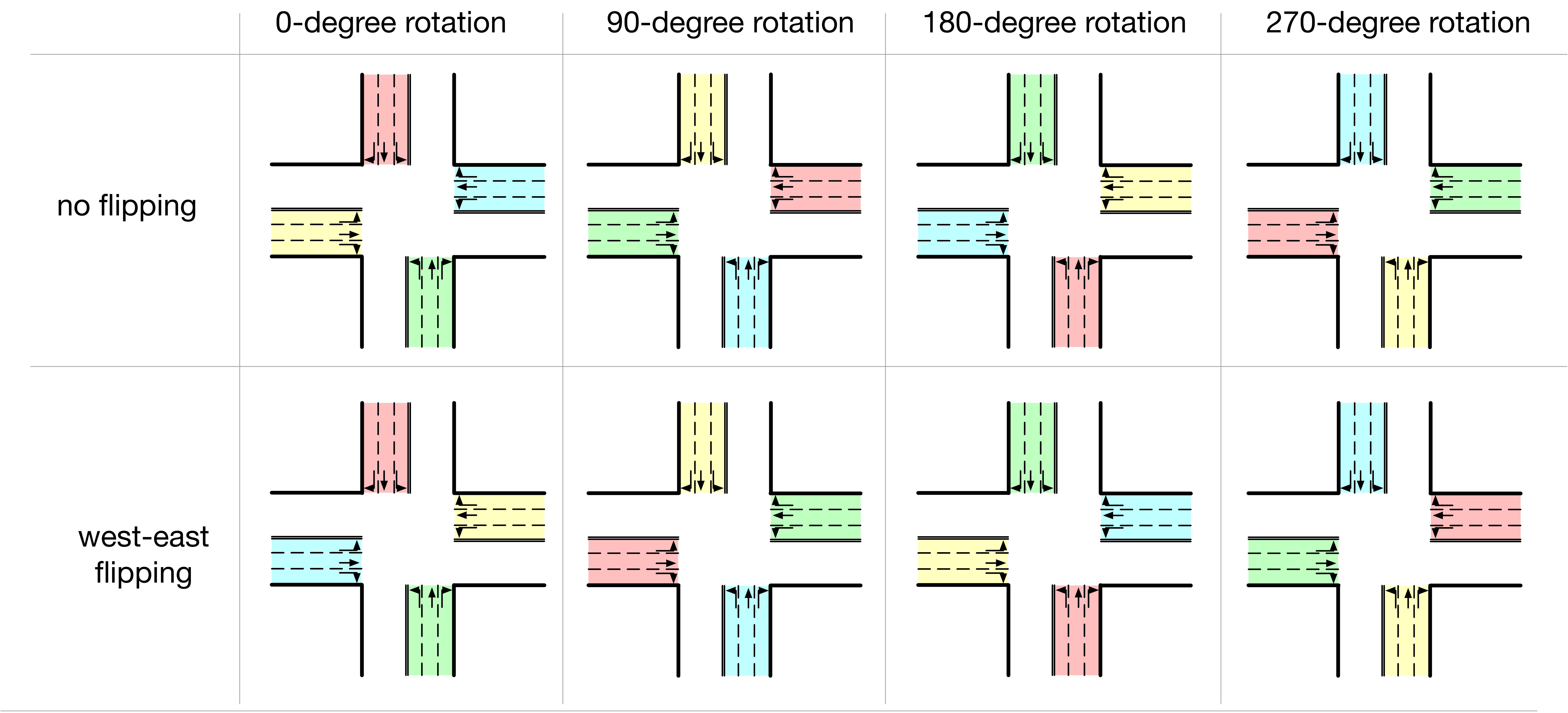}
\caption{All the variations based on rotation and flipping of the left-most case. Ideally, a RL model should handle all these cases equally well.}
\label{fig:rotate-flip}
\vspace{-4mm}
\end{figure}

Surprisingly, none of existing studies has attempted to address this issue. In fact, current RL methods are all exploring ``blindly'', wasting time on repeated situations. It is known that the principle of deep Q-network (DQN) is to use deep neural networks to approximate the state-action value $Q(s, a)$ and choose the action with the best value. Merely using fully-connected layers, previous RL methods such as \NIPS~\cite{VaOl16} and \deeplight~\cite{wei2018intellilight} regress the $Q(s, a)$ value for each phase from the 8-lane input independently, i.e., they have to go through roughly $8 \times n^8 \times8 $ samples for satisfactory approximation. But in fact, a considerable portion of state-action pairs are unnecessary to explore. Take Figure~\ref{fig:rotate} for example. These two scenarios are the same except that the traffic flow is flipped. Since such a flipping will result in a totally different state representation for existing methods, a RL agent which has learned the first case still cannot handle the second case. But based on the common sense, these two cases are almost identical and one would hope that the model learned from the first case can handle the second case or other similar cases. Furthermore, as shown in Figure~\ref{fig:rotate-flip}, given any particular state, one can generate seven other cases through rotation and flipping. An ideal RL model is thus expected to handle all eight cases even only one case is seen during training.


Based on the above observation, we propose a novel RL model design called \ours (which is invariant to symmetric operations like \textbf{F}lip and \textbf{R}otation and considers \textbf{A}ll \textbf{P}hase configurations). The key idea is that, instead of considering individual traffic movements, one should focus on the relative relation between different traffic movements. This idea is based on the intuitive principle of competition in traffic signal control: (1) larger traffic movement indicates higher demand for green signal; and (2) when two signals conflict, we should give the priority to the one with higher demand. 

Inspired by this principle, \ours first predicts the demand for each signal phase, and then models the competition between phases. Through the pair-wise phase competition modeling, \ours is able to achieve invariance to symmetries in traffic signal control (e.g., flipping and rotation). By leveraging such invariance and enabling knowledge sharing across the symmetric states, \ours successfully reduces the exploration space to $16\times n^4$ samples from $64 \times n^8$ (see Section~\ref{sec:method} for detailed analysis). Compared to existing RL-based methods, \ours finds better policies and converges much faster under complex traffic control scenarios. 

In summary, the main contributions of this paper include:
\begin{itemize}
    \item We propose a novel model design \ours for RL-based traffic signal control. By capturing the competition relation between different traffic movements, \ours achieves invariance to symmetry properties, which in turn leads to better solutions for the difficult all-phase traffic signal control problem.
    \item We demonstrate that \ours converges much faster than existing RL methods during the learning process through comprehensive experiments on real world data.
    \item We further demonstrate the superior generalizability of \ours. Specifically, we show that \ours can handle different road structures, different traffic flows, complex real-world phase settings, as well as a multi-intersection environment.
\end{itemize}

%% file: related_work.tex

\section{Related Work}
\label{sec:related-work}

\begin{figure*}[t]
\centering
\includegraphics[width=0.85\textwidth]{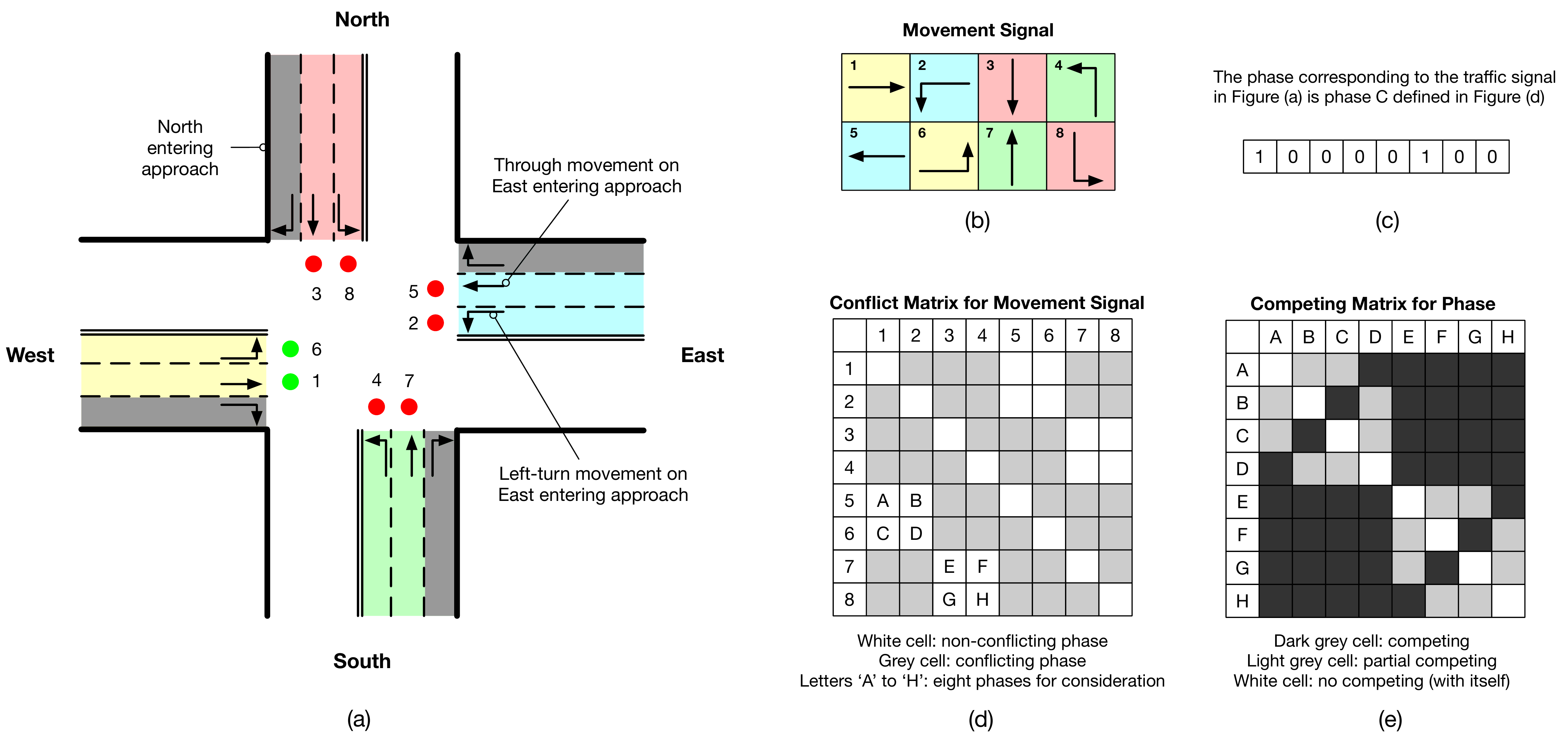}\vspace{-3mm}
\caption{Illustration of preliminary definition.}\vspace{-1mm}
\label{fig:phase}
\end{figure*}

\smallskip
\noindent\textbf{Traditional traffic signal control.}
Traffic signal control is a core research topic in transportation field and existing methods can be generally categorized into four classes.

\textit{Fixed-timed control}~\cite{Roess2011t} decides a traffic signal plan according to human prior knowledge and the signal timing does not change according to the real-time data. 

\textit{Actuated methods}~\cite{mirchandani2001real, fellendorf1994vissim} define a set of rules and the traffic signal is triggered according to the pre-defined rules and real-time data.  An example rule can be, to set the green signal for that traffic movement if the queue length is longer than certain threshold. 

\textit{Selection-based adaptive control methods} first decide a set of traffic signal plans and choose one that is the best for the current traffic situation (based on traffic volume data received from loop sensors). This method is widely deployed in today's traffic signal control. Commonly used systems include SCATS~\cite{SCATS,scats90}, RHODES~\cite{mirchandani2005rhodes} and SCOOT~\cite{hunt1982scoot}. 

All the methods mentioned above highly rely on human knowledge, as they require manually designed traffic signal plans or rules. 

\textit{Optimization-based adaptive control approaches} rely less on human knowledge and decide the traffic signal plans according to the observed data. These approaches typically formulate traffic signal control as an optimization problem under certain traffic flow model. To make the optimization problems tractable, strong assumptions about the model are often made. For example, a classical approach is to optimize travel time by assuming uniform arrival rate~\cite{webster1966traffic, Roess2011t}. The traffic signal plan including cycle length and phase ratio can then be calculated using a formula based on the traffic data. However, the model assumptions (e.g., uniform arrival rate~\cite{webster1966traffic, Roess2011t}) are often too restricted and do not apply in the real world. 

\smallskip
\noindent\textbf{Learning for traffic signal control.}
Different from traditional methods, learning-based traffic signal control does not require any pre-defined traffic signal plan or traffic flow models. In particular, reinforcement learning methods directly learns from intersections with the world. In these methods, each intersection is an agent, state is the quantitative description of the traffic condition at that intersection, action is the traffic signal, and reward is a measure on the transportation efficiency.

Existing RL methods differ in terms of state description of the environment (e.g., image of vehicle positions~\cite{KWBV08, BSSN+05, VaOl16,wei2018intellilight, XXL13}, queue length~\cite{APG03, AMB11, AMB14,wei2018intellilight, XXL13}, waiting time~\cite{Wier00,PBTB+13,BPT14,wei2018intellilight}), action definition (e.g., change to next phase~\cite{VaOl16,wei2018intellilight,PBTB+13,BPT14}, setting a phase~\cite{AMB11, AMB14,Saca10,SCGC08}), and reward design (e.g.,  queue length~\cite{MaDH16,VaOl16,BSSN+05}, delay~\cite{BPT14,ElAb10, ElAb12,VaOl16}).
In terms of algorithms, studies have utilized tabular methods (e.g., Q-learning~\cite{APG03,ElAb10}) for discrete state space and approximation methods~\cite{wei2018intellilight, liang2018deep}, which can be further categorized into value based (e.g., deep Q-Network~\cite{VaOl16, liang2018deep, wei2018intellilight}), policy based (e.g., policy gradient~\cite{mousavi2017traffic}), and actor critic~\cite{aslani2017adaptive, casas2017deep, aslani2018developing}.

However, to the best of our knowledge, none of these methods have shown satisfactory results in complete 8-phase scenario for one single intersection due to the large exploration space. In this paper, we follow the universal principles of competition and invariance in traffic signal control to design a novel model for efficient exploration. Further, we adopt the distributed framework of Ape-X DQN~\cite{apex} as our base framework, which is shown to achieve the state-of-the-art performance in playing Atari games. But our model design can be adapted to other algorithmic frameworks including policy based and actor critic based RL methods.

%% file: problem_definition.tex

\section{Problem Definition}
\label{sec:problem-definition}

\subsection{Preliminary}
\label{sec:problem-definition-problem-scenario}

In this paper, we investigate the traffic signal control in the scenario of a single intersection. To illustrate the definitions, we use the 4-approach intersection shown in Figure~\ref{fig:phase} as an example. But the concepts can be easily generalized to different intersection structures (e.g., different number of entering approaches). 
\begin{itemize}
    \item \textbf{Entering approach}: Each intersection has four entering approaches, named as North / South / West / East entering approach (`N', `S', `W', `E' for short) respectively.  In Figure~\ref{fig:phase}(a), we point out the North entering approach.  
    \item \textbf{Traffic movement}: A traffic movement is defined as the traffic moving towards certain direction, i.e., left turn, through, and right turn. In Figure~\ref{fig:phase}(a), we show that there are $8$ traffic movements. Follow the traffic rules in most countries, right turn traffic can pass regardless of the signal, but it needs to yield on a red light. In addition, a traffic movement could occupy more than one lane but this does not affect our model design because a traffic signal controls a traffic movement instead of a lane.
    \item \textbf{Movement signal}: For each traffic movement, we can use one bit with 1 as `green' signal and 0 as `red'.   
    \item \textbf{Phase}: We use an 8-bit vector $\mathbf{p}$ to represent a combination of movement signals (i.e., a phase), as shown in Figure~\ref{fig:phase}(b). As indicated by the conflict matrix in Figure~\ref{fig:phase}(d), some signals cannot turn `green' at the same time (e.g., signals \#1 and \#2). All the non-conflicting signals will generate 8 valid paired-signal phases (letters `A' to 'H' in Figure~\ref{fig:phase}(c)) and 8 single-signal phases (the diagonal cells in conflict matrix). Here we do not consider the single-signal phase  because in an isolated intersection, it is always more efficient to use paried-signal phases.\footnote{When considering multiple intersections, single-signal phase might be necessary because of the potential spill back.}
\end{itemize}

\subsection{RL Environment}

Driven by the idea of learning from the feedback, in this paper we propose a reinforcement learning approach to traffic signal control. In our problem, an agent can observe the traffic situation at an isolated intersection (Figure~\ref{fig:phase}(a)) and change the traffic signals accordingly. The goal of the agent is to learn a policy for operating the signals which optimizes travel time. This traffic signal control problem can be formulated as a Markov Decision Process $<\mathcal{S}, \mathcal{A}, \mathcal{P}, \mathcal{R}, \mathcal{\gamma}>$~\cite{SuBa98}:

\begin{problem}
Given the state observations set $\mathcal{S}$, action set $\mathcal{A}$, the reward function $\mathcal{R}$ is a function of $\mathcal{S \times A} \rightarrow \mathbb{R} $, specifically, $\mathcal{R}_s^{a} = \mathbb{E} [R_{t+1}|S_t = s, A_t = a]$. The agent aims to learn a policy $\pi(A_t = a|S_t = s)$, which determines the best action $a$ to take given state $s$, so that the following expected discounted return is maximized:\footnote{State transition probability matrix $\mathcal{P}$ is not described here because it is not explicitly modeled in model-free methods.}
\begin{equation}
    G_t = R_{t+1} + \gamma R_{t+2} + \gamma^2 R_{t+3} + ... = \sum_{m=0}^{\infty} \gamma_m R_{t+m+1}.
    \label{eq:return}
\end{equation}
\end{problem}

For traffic signal control, our RL agent is defined as follows:
\begin{itemize}
    \item {\bf State}: the number of vehicles $\mathbf{f}^v_i$ on each traffic movement $i$ and current traffic signal phase (represented as one bit $\mathbf{f}^s_i$ for each traffic movement signal).
    \item {\bf Action}: to choose the phase for the next time interval.
    \item {\bf Reward}: the average queue length of each traffic movement.
\end{itemize}

Note that, we use a relatively simple set of state features and reward, for the reason that we focus on innovating the model design in this paper. However, our method can easily incorporate more complex state features and rewards for performance boosts.

%% file: method.tex

\begin{figure*}[htbp]
\centering
\includegraphics[width=0.9\textwidth]{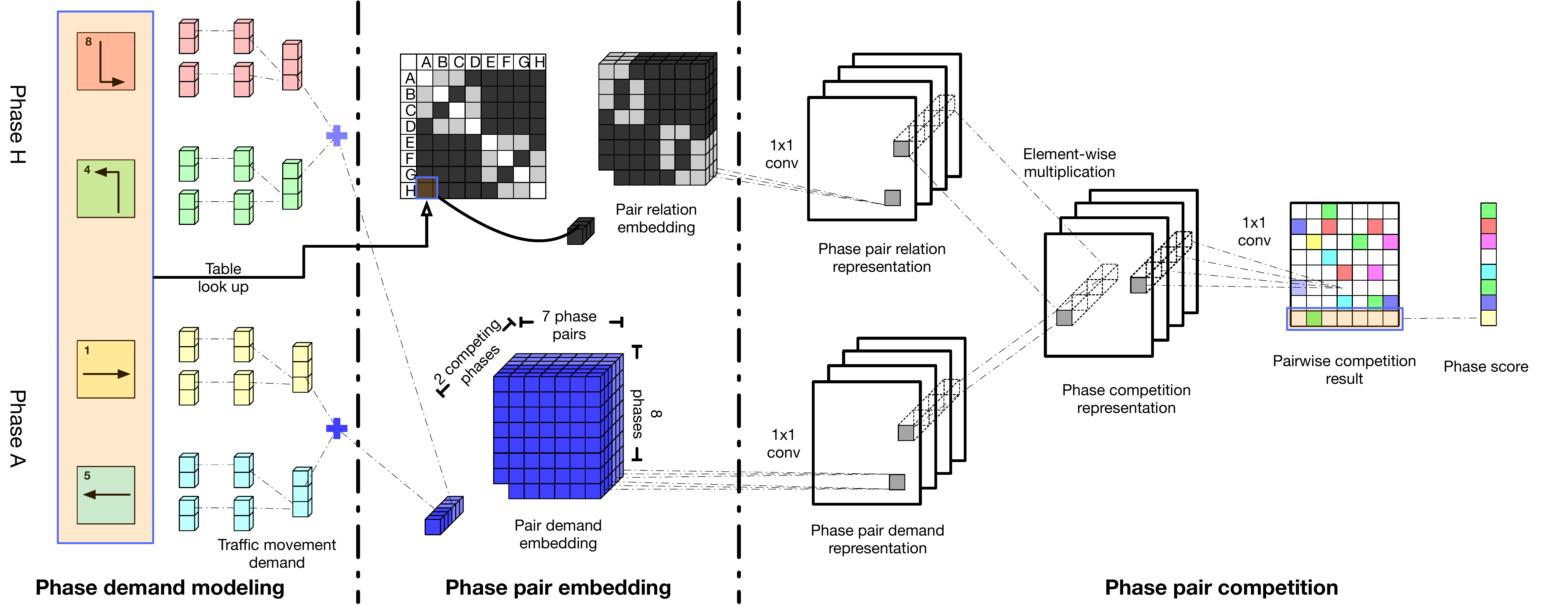}
\caption{Network design of \ours signal control.}
\label{fig:network}
\end{figure*}

\section{Method}
\label{sec:method}

\subsection{Model Overview}
\label{sec:method-overview}

 Similar to prior work~\cite{liang2018deep, wei2018intellilight}, we use Deep Q-learning (DQN) to solve the RL problem. Basically, our DQN network takes the state features on the traffic movements as input and predict the score (i.e., Q value) for each action (i.e., phase) as described in the Bellman Equation~\cite{SuBa98}:
\begin{equation}
Q(s_t,a_t)=R(s_t, a_t)+\gamma \max Q(s_{t+1}, a_{t+1}).
\label{eq:bellman}
\end{equation}
 
 We design a novel network called \ours (which is invariant to symmetric operations like \textbf{F}lip and \textbf{R}otation and considers \textbf{A}ll \textbf{P}hase configurations) based on two universal principles:
\begin{itemize}
    \item {\bf Principle of competition}: Larger traffic movement indicates higher demand for `green' movement signal. When two signals conflict, priority should be given to the one with higher demand.
    \item {\bf Principle of invariance}: Signal control should be invariant to symmetries such as rotation and flipping.
\end{itemize}

This way, the learning for different traffic movements and phases can now occur at the same time by updating the same network module (i.e., parameters), which leads to more efficient use of the data samples and better performance.
 
 The rest of this section is organized as follows. In Section~\ref{sec:algorithm-framework}, we give a brief overview of the state-of-the-art Ape-X DQN~\cite{apex} framework, upon which our method is built. Then, we describe our network design in details in Section~\ref{sec:method-network-design}. In Section~\ref{sec:method-fast-adaptation}, we further discuss some important properties of our model.

\subsection{Algorithmic Framework}
\label{sec:algorithm-framework}

For improved learning efficiency with the large search space, we adopt the distributed framework Ape-X DQN~\cite{apex} as our algorithmic framework. 
In Ape-X DQN, the standard deep reinforcement learning is decomposed into two parts: \textit{acting} and \textit{learning}. 
The \textit{acting} part assigns multiple actors with different exploration policies to interact with an environment and to store the observed date in a replay memory. The \textit{learning} part is responsible for sampling the training data in the replay memory to update the model. 
Most importantly, the two parts can run concurrently while keeping the speed of generating and consuming the training data almost equal. In short, benefiting from the high exploration and sampling efficiency, this framework can significantly boost the learning performance of reinforcement learning. For more details about Ape-X DQN, we refer interested readers to~\cite{apex}.

\subsection{Phase Invariant Signal Control Design}
\label{sec:method-network-design}

As we discussed before, training a RL agent for traffic signal control is highly challenging due to the large search space. For instance, for the four-approach intersection shown in Figure~\ref{fig:phase}(a), assuming there is only one lane on each traffic movement, the size of the state space will be $n^8 \times 8$, where $n$ is the capacity of a lane. Thus, even with a small lane capacity (e.g., $n=10$), DQN will require billions of data samples to learn the relation between state, action and reward. Further, intersections may vary in the geometry (e.g., 3, 4, or 5 entering approaches) and the signal setting (i.e., different combination of traffic movement signals). It is very inefficient if a different agent needs to be learned for each different intersection.

To address these challenges, we design our model based on the two principles outlined in Section~\ref{sec:method-overview}, so that it can learn more efficiently from data and also be easily adapted to different intersection structures. We divide the prediction of phase score (i.e., Q value) into three stages: \emph{phase demand modeling}, \emph{phase pair embedding}, and \emph{phase pair competition}. Figure~\ref{fig:network} shows an overview of our method. Next we describe these three stages in details.

\subsubsection{Phase Demand Modeling}
In this stage, our goal is to obtain a representation of the demand for each signal phase. Recall that for any traffic movement $i, i\in\{1,\ldots, 8\}$, its state includes the number of vehicles and the current signal phase. These features can be obtained directly from the simulator. We first take these two features, denoted as $\mathbf{f}^v_i$ and $\mathbf{f}^s_i$ respectively, as input, and pass them through a neural network of two fully-connected layers to generate a representation of the demand for `green' signal on this traffic movement, $\mathbf{d}_i$.
\begin{equation}
\mathbf{h}^v_i = ReLU(\mathbf{W}^v\mathbf{f}^v_i + \mathbf{b}^v), \quad
\mathbf{h}^s_i = ReLU(\mathbf{W}^s\mathbf{f}^s_i + \mathbf{b}^s).
\end{equation}
\begin{equation}
\mathbf{d}_i = ReLU(\mathbf{W}^h[\mathbf{h}^v_i, \mathbf{h}^s_i] + \mathbf{b}^h).
\end{equation}
Note that the two hidden layer vectors $\mathbf{h}^v_i, \mathbf{h}^s_i$ are combined before passed through the output layer. Further, the learned parameters of the neural network are shared among all traffic movements.

Finally, we obtain the demand representation of any phase $\p$ by adding together the demands of the two non-conflicting traffic movement signals in $\p$:
\begin{equation}
\mathbf{d}(\p) = \mathbf{d}_i + \mathbf{d}_j, \text{ where } \p_i=\p_j=1.
\end{equation}

\subsubsection{Phase Pair Embedding}
By the principle of competition, the score (priority) of a phase depends on its competition with the other phases. Thus, for each phase $\p$, we form phase pairs $(\p, \q)$ where $\q$ is an opponent of $\p$ (i.e., $\q \neq \p$). Given a pair $(\p, \q)$ and their demands, our goal of this stage is therefore to obtain a representation of the competition between $\p$ and $\q$. We observe that two aspects are essential to the competition: their relation and their demands. Consequently, we generate two embeddings to capture the aspects.

\smallskip
\noindent{\bf Pair relation embedding}: As illustrated in Figure~\ref{fig:phase}(e), a phase pair $\p$ and $\q$ can have two different relations: partial competing (light grey, e.g., phase A and B, which shares one traffic movement) and competing (dark grey, e.g., phase A and D, which totally conflict with each other). Once the phase pair is determined, (e.g., phase A and phase B), their conflict matrix can be directly represented by the corresponding cell in the matrix in Figure~\ref{fig:phase}(d). Our embedding model will look up the phase competing matrix and map the relation to an embedding vector $\mathbf{e}(\p,\q)$. Putting together the embedding vectors of all the phase pairs forms the relation embedding volume $\mathbf{E}$ (grey volume in Figure~\ref{fig:network}).

\smallskip
\noindent{\bf Pair demand embedding}:
Similarly, the pair demand embedding volume $\mathbf{D}$ (blue volume in Figure~\ref{fig:network}) is formed by first concatenating the demand representations of phases in a phase pair (i.e., $[\mathbf{d}(\p), \mathbf{d}(\q)]$), and then gathering the vectors of all phase pairs. 

For our 8-phase problem, the sizes of $\mathbf{E}$ and $\mathbf{D}$ are $8\times 7\times l_1$ and $8\times 7\times l_2$, where $l_1$ and $l_2$ are the length of the relation embedding vector and demand embedding vector for a single pair, respectively.

\subsubsection{Phase Pair Competition}

In this stage, our model takes the phase pair embedding volumes $\mathbf{E}$ and $\mathbf{D}$ as input and predicts the score (i.e., Q-value) of each phase considering its competition with other phases. 

We first process two volumes separately by feeding each of them into $K$ convolutional layers with $1 \times 1$ filters. The choice of $1 \times 1$ filters follows the idea of extracting competing relationship in the phase pair by letting them interacting with each other, which is verified in prior work~\cite{lin2013network}. $1 \times 1$ filter also enables the parameter sharing among different phase pairs. Because there are no explicit meaningful interaction between different phase pairs (e.g., 2, 3, 4, or more phase pairs), it is not useful to use larger filters. Then the $k$-th layer can be written as:
\begin{equation}
    \mathbf{H}^r_k = \mathrm{ReLU}(\mathbf{W}^r_k \cdot \mathbf{H}_{k-1}^r + \mathbf{b}^r_k),
\end{equation}
\begin{equation}
    \mathbf{H}^d_k = \mathrm{ReLU}(\mathbf{W}^d_k \cdot \mathbf{H}_{k-1}^r + \mathbf{b}^d_k),
\end{equation}
where $\mathbf{H}^r_0 = \mathbf{E}$, and $\mathbf{H}^d_0 = \mathbf{D}$.

After that, a phase competition representation $\mathbf{H^c}$ can be obtained by an element-wise multiplication of the phase pair demand representation $\mathbf{H}^d_K$ and the phase pair relation representation $\mathbf{H}^r_K$: $\mathbf{H}^c = \mathbf{H}^d_K \otimes \mathbf{H}^r_K$. We then apply another convolutional layer with $1 \times 1$ filter to get the pairwise competition result matrix $\mathbf{C}$, each row of which represents the relative priorities of a phase $\p$ over all its opponents. Mathematically, we have 
\begin{equation}
    \mathbf{C} = \mathrm{ReLU}(\mathbf{W}^c \cdot \mathbf{H}^c + \mathbf{b}^c).
\end{equation}
Finally, the relative priorities of each phase $\p$ are added together to obtain the score of phase $\p$. Our RL agent then choose the phase with the highest score as its action.

\subsection{Discussions}
\label{sec:method-fast-adaptation}

\noindent{\bf Invariance of our model design.} Throughout the above modeling process, no matter which phase $\p$ we are focusing on, we always have a symmetric view of its relation with other phases. This enables \ours to leverage the symmetry properties in traffic signal control and greatly reduce the exploration of samples. Specifically, assume that a maximum of $n$ vehicles is allowed on each movement. Note that, the traffic movement signal could be either `1' (green) or `0' (red). As shown in Figure~\ref{fig:network}, our network first obtains a representation for a phase from features of two traffic movements with $2^2 \times n^2$ possible combinations. Then two phases are paired together to compete and the model of phase competition is shared among all pairs. In this way, to regress $Q$ values for all eight actions, the model is required to observe only $(2^2\times n^2)\times (2^2\times n^2)=16 \times n^4$ samples, a significant decrease in comparison with $64\times n^8$ as in \NIPS~\cite{VaOl16} and \deeplight~\cite{wei2018intellilight}. In Section~\ref{sec:experiment}, we further conduct extensive experiments to illustrate that \ours converges faster and to better solutions, as the enhanced sample efficiency compensates for the explosion of state space.

\begin{figure}[t]
\centering
\begin{tabular}{ccc}
\includegraphics[width=0.14\textwidth]{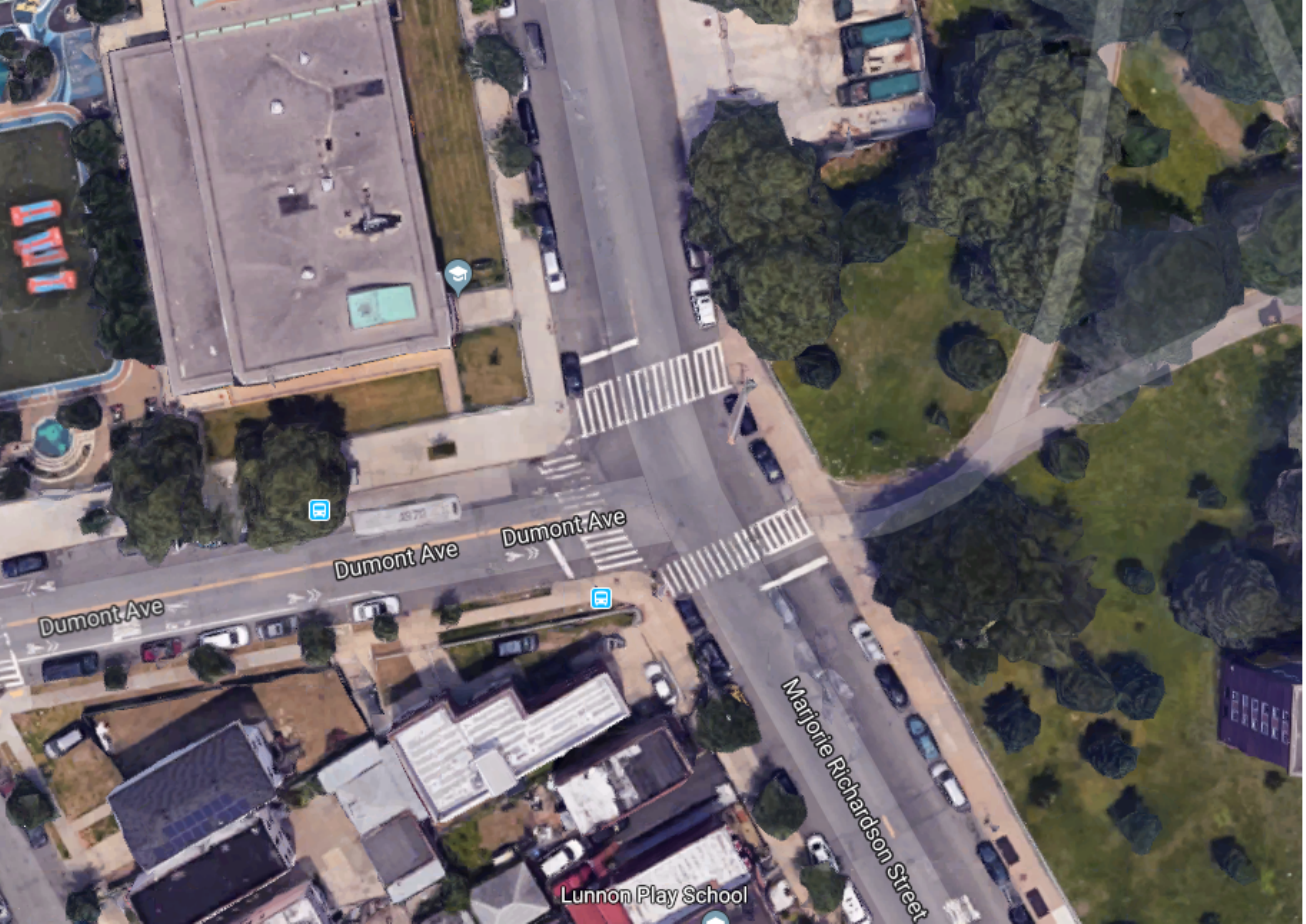}
& \includegraphics[width=0.14\textwidth]{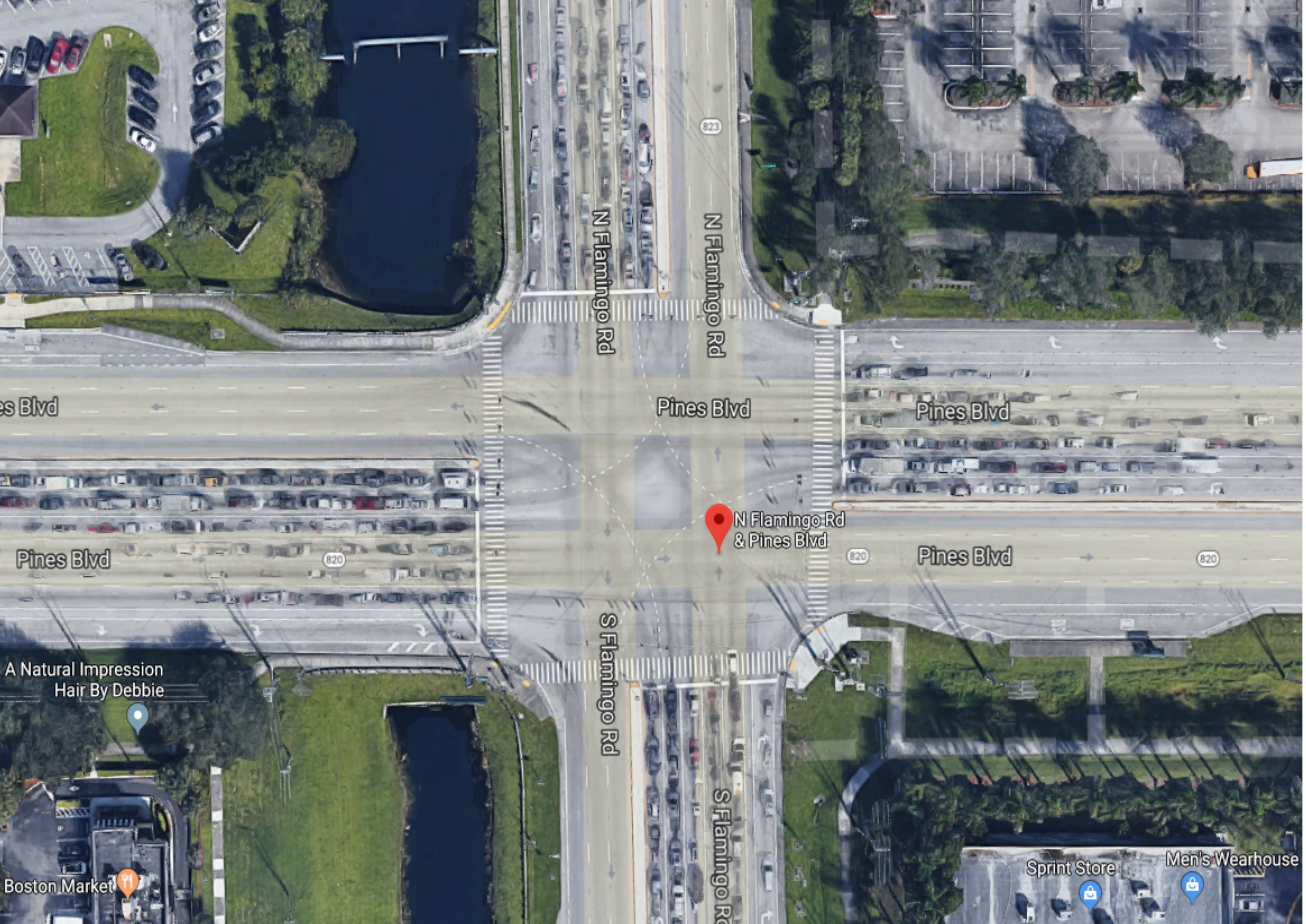} 
& \includegraphics[width=0.14\textwidth]{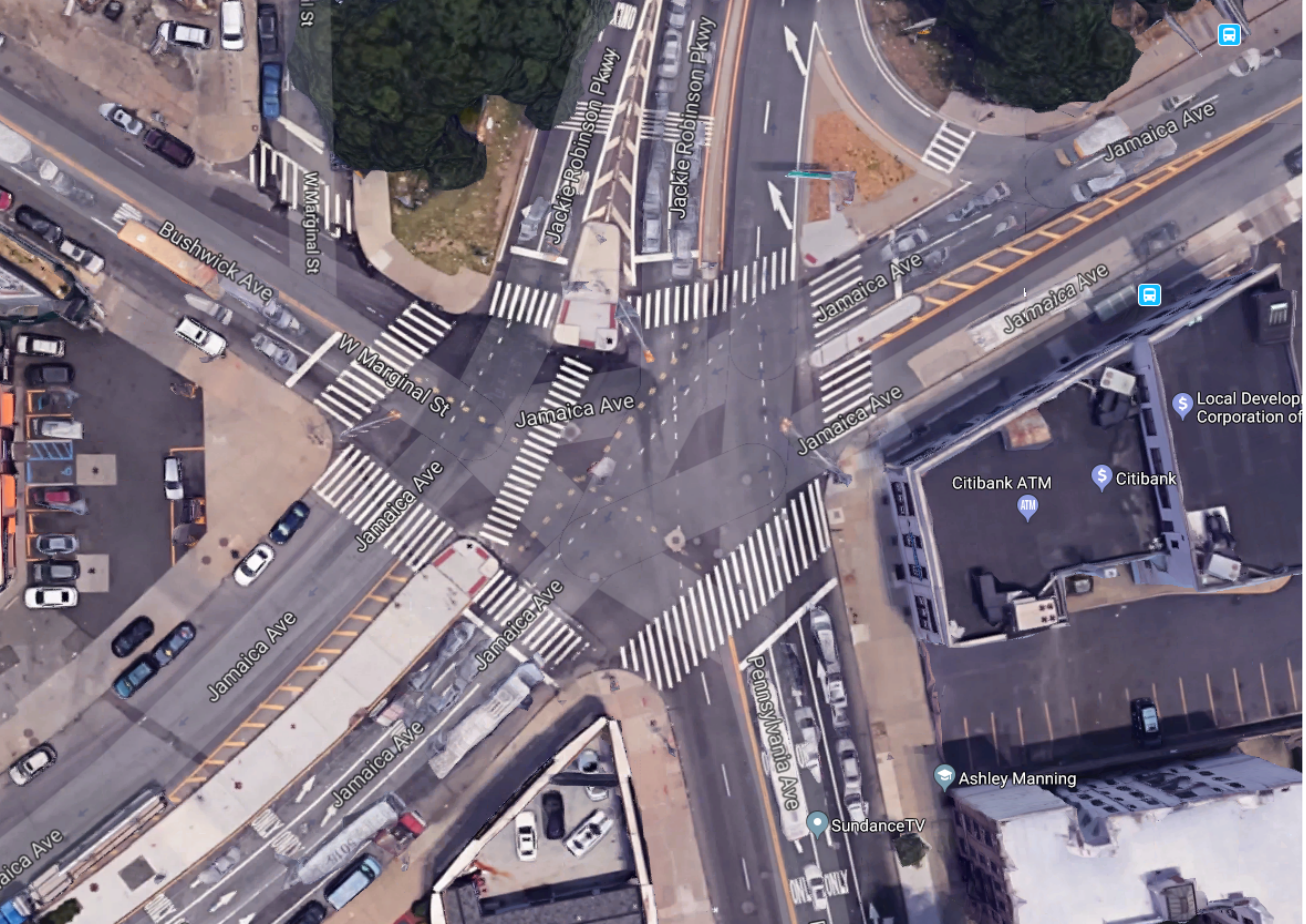}\\ 
\end{tabular}
\vspace{-3mm}
\caption{Real world 3-, 4-, and 5-approach intersections.}
\label{fig:complex-real-inter}
\end{figure}

\smallskip
\noindent{\bf Adaption to different environments.}
As we focus on the universal principles of phase competition and invariance throughout our model design, the \ours model can be applied to different traffic conditions (i.e., small, medium and large traffic volumes), different traffic signal settings (e.g., 4-phase, 8-phase), and different road structures (e.g., 3-, 4-, and 5-approach intersections as shown in Figure~\ref{fig:complex-real-inter}, and intersections with variable number of lanes on each traffic movement). Further, \ours can learn from one environment and transfer to another one with high accuracy without any additional training. We further illustrate this in the experiments.

\smallskip
\noindent{\bf Applications in multi-intersection environments.} Though our discussion so far has been only focused on single intersections, \ours makes fundamental contributions to city-wide traffic signal control, as a good learning model at single intersection is the base unit even in the scale of city-wide traffic signal control. In addition, we demonstrate that \ours works well in the multi-intersection environment even without explicit coordination (see experiments).

%% file: experiment.tex

\begin{table*}[tbhp]
\centering
\caption{Overall performance. Travel time is reported in the unit of second.}
\label{tab:overall}
\begin{tabular}{c|ccccccc|cccccc}
\toprule
\multirow{2}{*}{Model} & \multicolumn{7}{c|}{Jinan} & \multicolumn{6}{c}{Hangzhou} \\ 
& 1  & 2  & 3 & 4 & 5 & 6 & 7 & 1 & 2 & 3 & 4 & 5 & 6 \\
\midrule
\fixedtime & 118.82 & 250.00 & 233.83 & 297.23 & 101.06 & 104.00 & 146.66 & 271.16 & 192.32 & 258.93 & 207.73 & 259.88 & 237.77  \\
\formula & 107.92 & 195.89 & 245.94 & 159.11 & 76.16 & 100.56 & 130.72 & 218.68 & 203.17 & 227.85 & 155.09 & 218.66 & 230.49  \\
\SOTL & 97.80 & 149.29 & 172.99 & 64.67 & 76.53 & 92.14 & 109.35 & 179.90 & 134.92 & 172.33 & 119.70 & 188.40 & 171.77 \\
\hline
\NIPS & 98.90 & 235.78 & 182.31 & 73.79 & 66.40 & 76.88 & 119.22 & 146.50 & 118.90 & 218.41 & 80.13 & 120.88 & 147.80\\
\deeplight & 88.74 & 195.71 & 100.39 & 73.24 & 61.26 & 76.96 & 112.36 & 97.87 & 129.02 & 186.04 & 81.48 & 177.30 & 130.40  \\
\hline
\ours & \textbf{66.40} & \textbf{88.40} & \textbf{84.32} & \textbf{33.83} & \textbf{54.43} & \textbf{61.72} & \textbf{72.31} & \textbf{80.24} & \textbf{79.43} & \textbf{110.33} & \textbf{67.87} & \textbf{92.90} & \textbf{88.28} 
\\
\hline
Improvement & 25.17\% & 40.79\% & 16.01\% & 47.69\% & 11.15\% & 19.72\% & 33.87\% & 18.01\% & 33.20\% & 35.98\% & 15.30\% & 23.15\% & 32.30\% \\
\bottomrule
\end{tabular}
\label{table:overall}
\end{table*}

\section{Experiment}
\label{sec:experiment}

\subsection{Experiment Settings}

Following the tradition of the traffic signal control study~\cite{wei2018intellilight}, we conduct experiments in a simulation platform SUMO (Simulation of Urban MObility)\footnote{\url{http://sumo.dlr.de/index.html}}. After the traffic data being fed into the simulator, a vehicle moves to its destination according to the setting of the environment. The simulator provides the state to the signal control method and executes the traffic signal actions from the control method. Following the tradition, each green signal is followed by a three-second yellow signal and two-second all red time. 

In a traffic dataset, each vehicle is described as $(o, t, d)$, where $o$ is  origin location, $t$ is  time, and $d$ is  destination location. Locations $o$ and $d$ are both locations on the road network. Traffic data is taken as input for simulator.

In a multi-intersection network setting, we use the real road network to define the network in simulator. For a single intersection, unless otherwise specified, the road network is set to be a four-way intersection, with four 300-meter long road segments. 

\subsection{Datasets}
We use two private real-world datasets from Jinan and Hangzhou in China and one public dataset from Atlanta in the United States.

\noindent{\bf Jinan.}
We collect data from our collaborators in Jinan from surveillance cameras near intersections. There are in total 7 intersections with relatively complete camera records for single intersection control. Each record in this dataset contains the camera location, the time when one vehicle arrived at the intersection, and the vehicle information. These records are recovered from the camera recordings by advanced computer techniques. We feed the vehicles to the intersections at their recorded arrival time in our experiments. 

\noindent{\bf Hangzhou.}
This dataset is captured by surveillance cameras in Hangzhou from 04/01/2018 to 04/30/2018. There are in total 6 intersections with relatively complete camera records. These records are processed similarly as the Jinan data. 

\noindent{\bf Atlanta.}
This public dataset\footnote{\url{https://ops.fhwa.dot.gov/trafficanalysistools/ngsim.htm}} is collected by eight video cameras from an arterial segment on Peachtree Street in Atlanta, GA, on November 8, 2006. This vehicle trajectory dataset provides the precise location of each vehicle within the study area and five intersections in total are taken into consideration.

\begin{table}[tbh]
\centering
\caption{Overall performance.}
\label{tab:GA}
\begin{tabular}{c|ccccccc}
\toprule
\multirow{2}{*}{Model}&  \multicolumn{5}{c}{Atlanta} \\ 
     & 1 & 2 & 3 & 4 & 5\\
\midrule
\fixedtime & 140.51 & 334.17 & 334.01 & 353.56 & 271.14 \\
\formula & 116.16 & 148.71 & 163.93 & 157.08 & 254 \\
\SOTL & 101.87 & 133.79 & 136.77 & 138.75 & 73.93 \\
\hline
\NIPS & 152.93 & 95.83 & 101.61 & 79.03 & 43.04 \\
\deeplight & 76.25 & 74.10 & 83.12 & 65.94 & 47.51 \\

\hline 
\ours & \textbf{67.45} & \textbf{61.48} & \textbf{65.26} &  \textbf{60.75} & \textbf{41.39}\\
\hline
Improvement & 11.54\% & 17.03\% & 21.49\% & 7.87\% & 3.83\%\\
\bottomrule
\end{tabular}
\vspace{-2mm}
\end{table}

\subsection{Methods for Comparison}

To evaluate the effectiveness and efficiency of our model, we compare it with the following classic and state-of-the-art methods. We tune the parameters of each method separately and report the best performance obtained.

\begin{itemize}
    \item \textbf{\fixedtime}~\cite{Miller1963}: Fixed-time control adopts a pre-determined cycle and phase time plan, which is widely used in the steady traffic flow. A grid search is conducted to find the best cycle.
    \item \textbf{\SOTL}~\cite{cools2013self}: Self-Organizing Traffic Light Control is an approach which can adaptively regulate traffic lights based on a hand-tuned threshold on the number of waiting vehicles. 
    \item \textbf{\formula}: This method computes a reasonable cycle length of the traffic signal from the traffic condition, i.e., the preset volume for a uniform flow. Then the time assigned to each phase is decided by the traffic volume ratio.
    \item \textbf{\NIPS}~\cite{VaOl16}: This method leverages a DQN framework for traffic light control and takes as state an image depicting vehicles' positions on the road. 
    \item \textbf{\deeplight}~\cite{wei2018intellilight}: This is another deep reinforcement learning method with a more elaborate network architecture. This is the state-of-the-art RL method and demonstrates good performance in 2-phase signal control.
\end{itemize}

\subsection{Evaluation Metrics}
Based on existing studies in traffic signal control, we choose a representative metric, \textbf{travel time}, for evaluation. This metric is defined as average travel time vehicles spend on approaching lanes (in seconds), which is the most frequently used measure to judge performance in the transportation field.

\subsection{Overall Performance}

Table~\ref{tab:overall} and \ref{tab:GA} report the travel times achieved by all methods with the 8-phase setting. Note that \textbf{Improvement} is the percentage by which \ours surpasses the best baseline. We can see clearly that our method significantly outperforms all other methods on all datasets. 

\begin{figure}[htbp]
\centering
\includegraphics[width=0.36\textwidth]{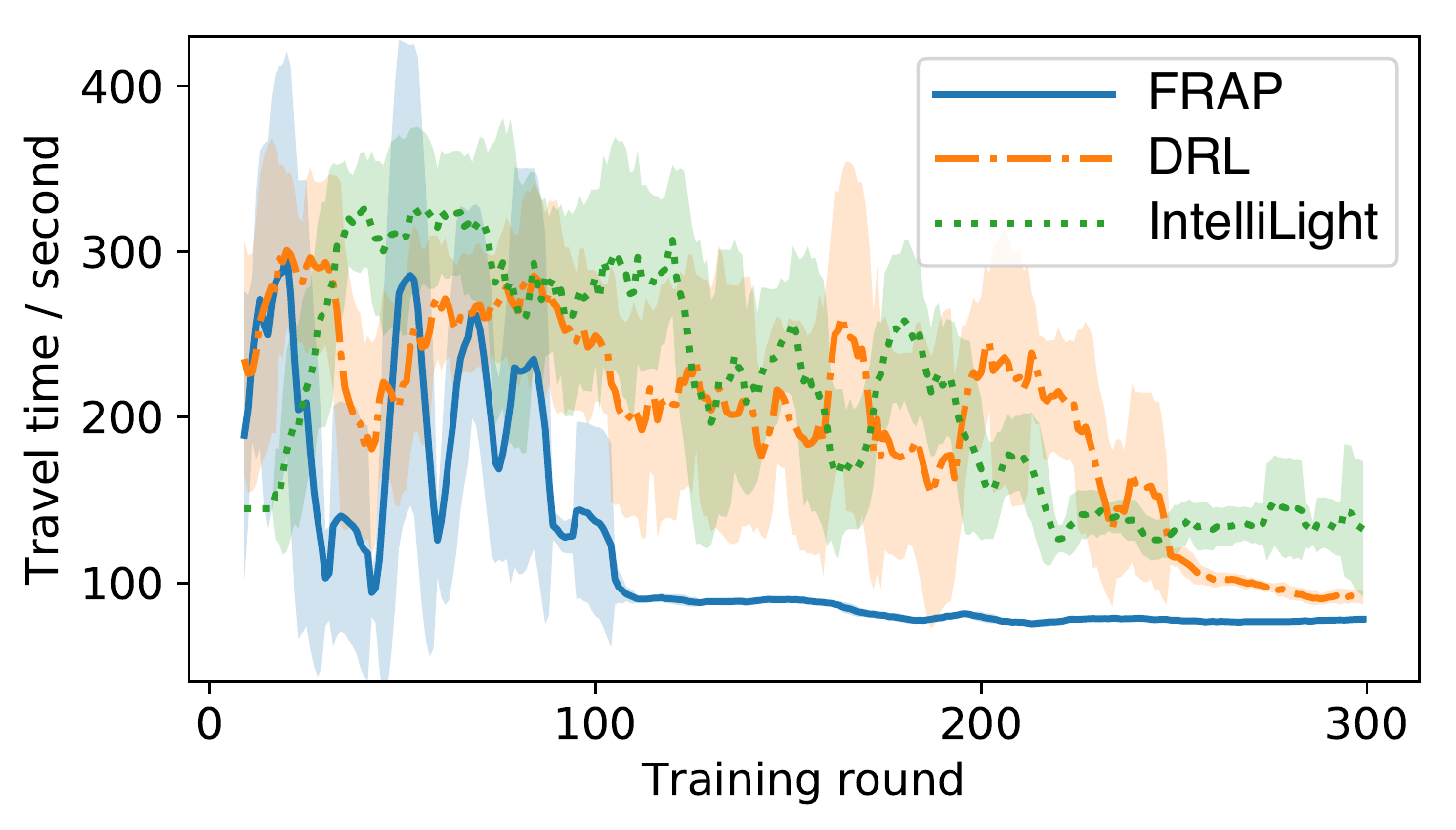} \vspace{-2mm}
\caption{Convergence speed of RL methods.}\vspace{-2mm}
\label{fig:overall-convergence}
\end{figure}

As expected, RL methods tend to perform better than conventional ones like \fixedtime as the ability to capture real-time information at the intersection enables RL methods to make more reasonable decisions. Among these RL approaches, our method stands out not only in terms of travel time, but also in terms of convergence speed. Figure~\ref{fig:overall-convergence} plots convergence curves of RL methods and \ours leads to the fastest convergence (we only show one case due to space limit). It is because \ours leverages the symmetry properties of traffic signal control and the Ape-X DQN framework to improve sampling efficiency.

\subsection{Model Characteristics}
\subsubsection{Invariance to flipping \& rotation}
\begin{figure}[htbp]
\centering
\includegraphics[width=0.38\textwidth]{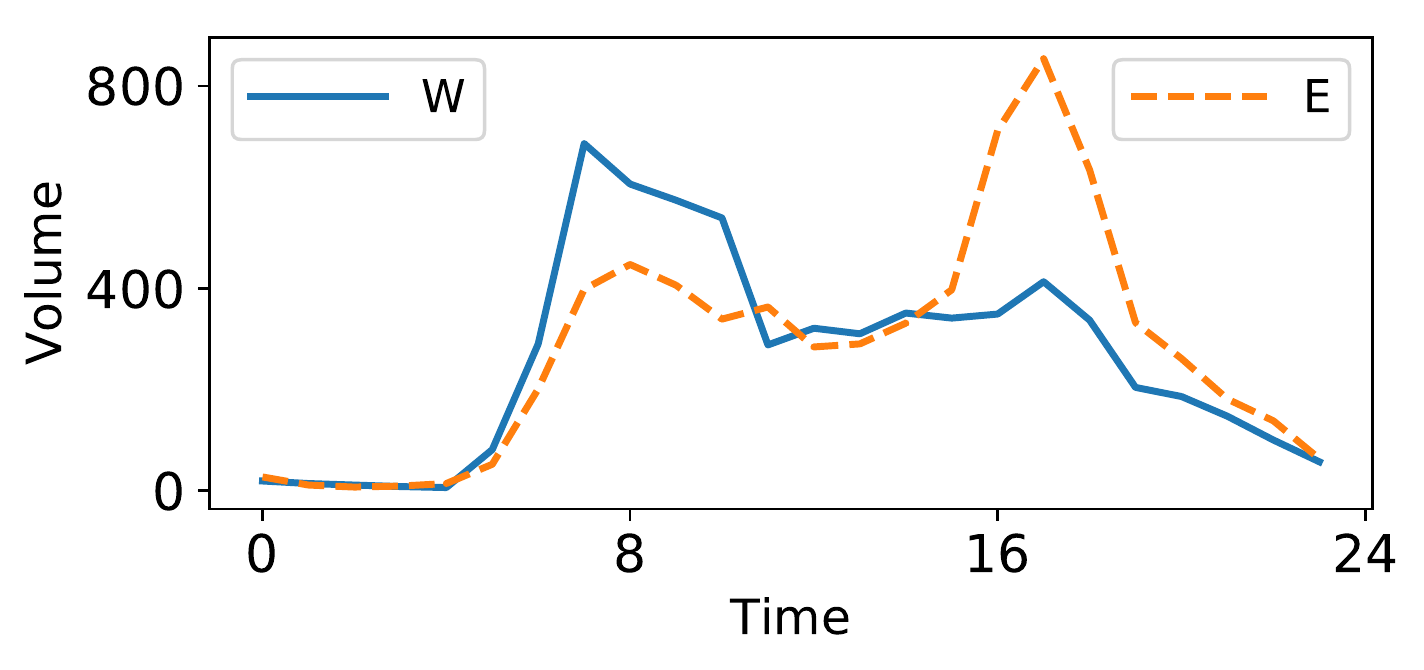}
\vspace{-2mm}
\caption{An instance of traffic flow flipping from the intersection 4 in Jinan, China.}\vspace{-2mm}
\label{fig:traffic-flip}
\end{figure}

Besides achieving faster travel time and convergence speed, \ours has another advantage in its invariance to flipping and rotation. In the real world, it is common that people drive to work in a specific movement in the morning and go home in the opposite direction in the afternoon. Figure~~\ref{fig:traffic-flip} shows an example traffic flow flipping from intersection 4 in Jinan. It can be observed that the traffic volume of the west approach is much larger than that of the east approach at around 8 am.  At 5 pm, the relation is reversed.

\begin{figure}[hbt]
  \centering
  \begin{tabular}{cc}
   \hspace{-2mm}\includegraphics[width=0.235\textwidth]{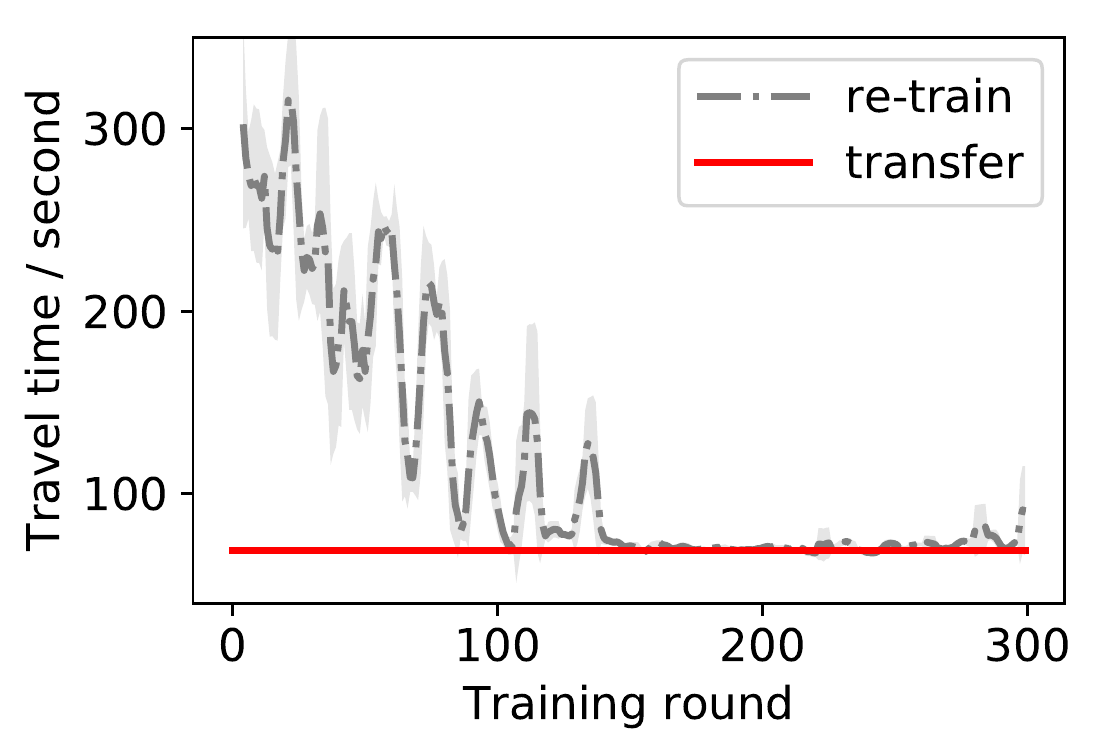}&
   \hspace{-2mm}\includegraphics[width=0.235\textwidth]{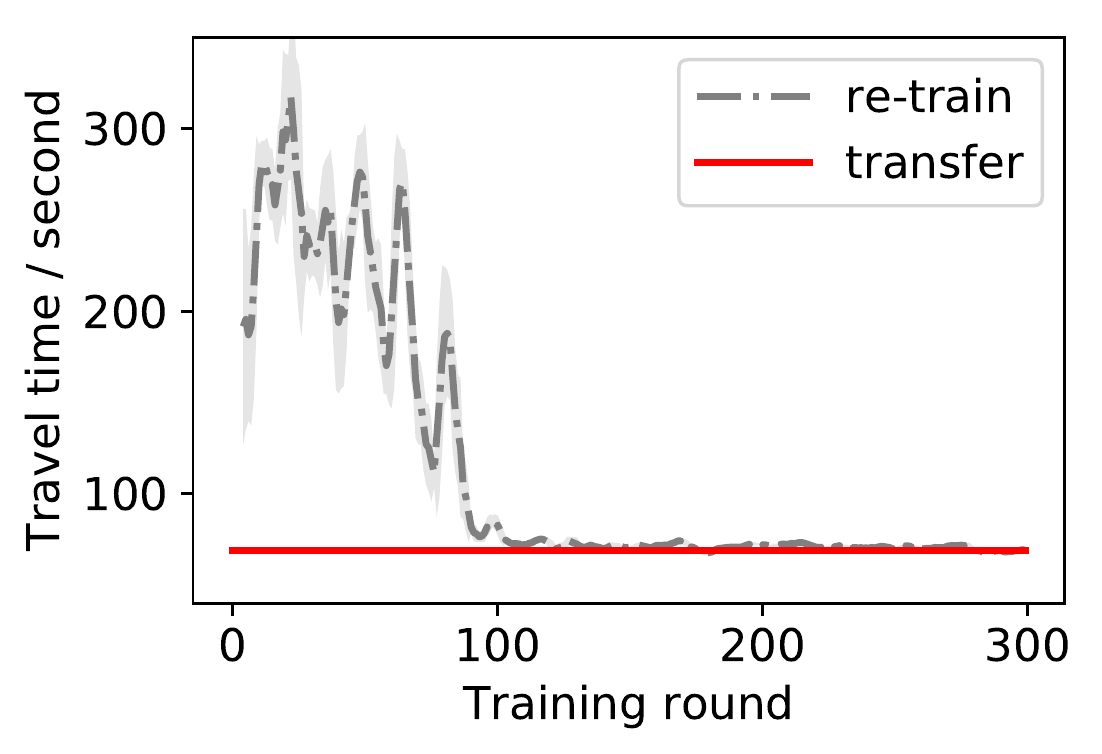} \\
   (a) Flipping & (b) Rotation\\
   \end{tabular}
 \caption{Invariance to flipping and rotation. We show the convergence curves of re-trained and transferred models for flipped and rotated traffic flow.}
    \label{fig:flip-rotate}
\end{figure}

Without using the symmetry properties of traffic signal control, previous RL methods have to re-train a model when the flow varies drastically (e.g., flipping and rotation). However, the \ours model design guarantees that our method is less vulnerable to such extreme changes, meaning that the model will perform nearly the same under those traffic flows. Figure~\ref{fig:flip-rotate} shows \ours's invariance to flipping and rotation. In this experiment, we directly take a model trained from the original traffic flow and test it on the flipped and rotated flows. We compare its performance with two re-trained models. From this figure, we can observe that our transferred model achieves almost identical travel time performance to the re-trained models, thus spares the extra training costs.

\subsubsection{Adaptation to different traffic volumes} In this experiment, we illustrate another advantage of \ours over other RL methods in its adaptability to traffic volume. Intuitively, if a model explores sufficient states when trained on a heavy traffic flow, it can adapt to those relatively light flows. However, due to the large state space, existing RL methods can hardly see enough samples for the transfer to different traffic flows. In the meantime, the \ours model design takes advantages of the symmetry propoerties of traffic signal control to improve data efficiency, which leads to better transferability.

\begin{figure}[htbp]
\centering
\includegraphics[width=0.38\textwidth]{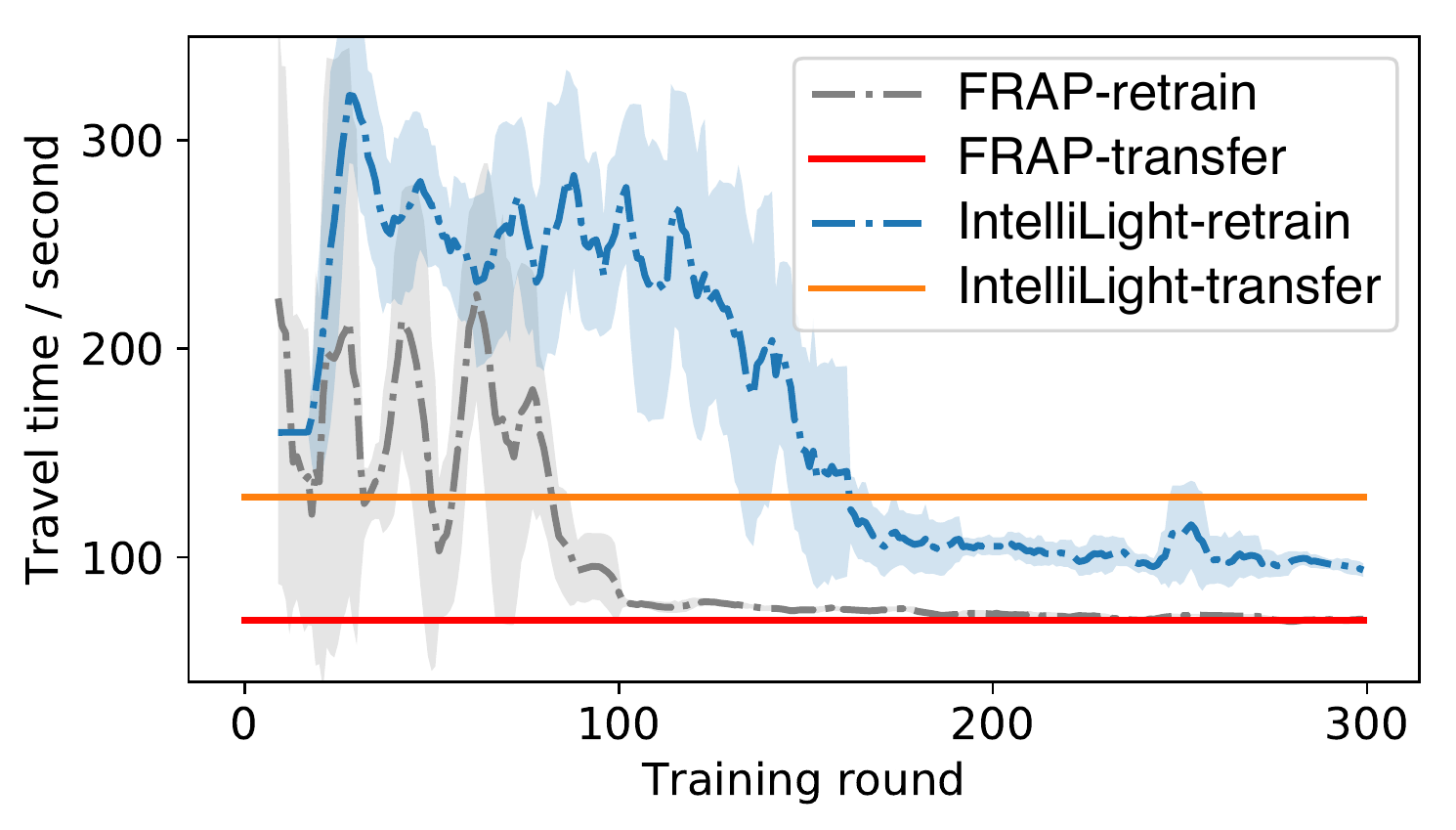}\vspace{-2mm}
\caption{\ours model design leads to higher data efficiency and better adaptation to different traffic volumes.}\vspace{-2mm}
\label{fig:traffic-adaptation}
\end{figure}

In this experiment, we choose both \ours and \deeplight models trained from the intersection 2 in Jinan with the largest vehicle volume, and evaluate their performances on a relatively light flow from the intersection 1. In this case, both models have explored similar states and converged to the best values they can achieve for intersection 2. From Figure~\ref{fig:traffic-adaptation}, we can see that the transferred model of \ours performs almost the same as the re-trained model, whereas there is a distinct gap between re-trained and transferred models of \deeplight. This suggests that the proposed \ours model design increases sampling efficiency significantly and facilitates adaptation to different traffic flows.

\subsubsection{Flexibility of the 8-phase setting}
Due to transportation traditions, an intersection with the 4-phase setting (Phase A, D, E, and H) is also very common in the real world. Although the 4-phase setting is simple and efficient sometimes, it has a severe limitation in that the green time for through or left-turn signal is always the same in the two opposite approaches. This will exert negative effects on travel time when the volume is unbalanced in these two approaches, which occurs frequently on real roads. Meanwhile, the 8-phase setting allows vehicles from one approach to pass exclusively. In this experiment, we show that the 8-phase setting is more flexible than the 4-phase setting, that is, under the 8-phase setting vehicles can pass faster and more reasonably.

\begin{figure}[htbp]
\centering
\includegraphics[width=0.38\textwidth]{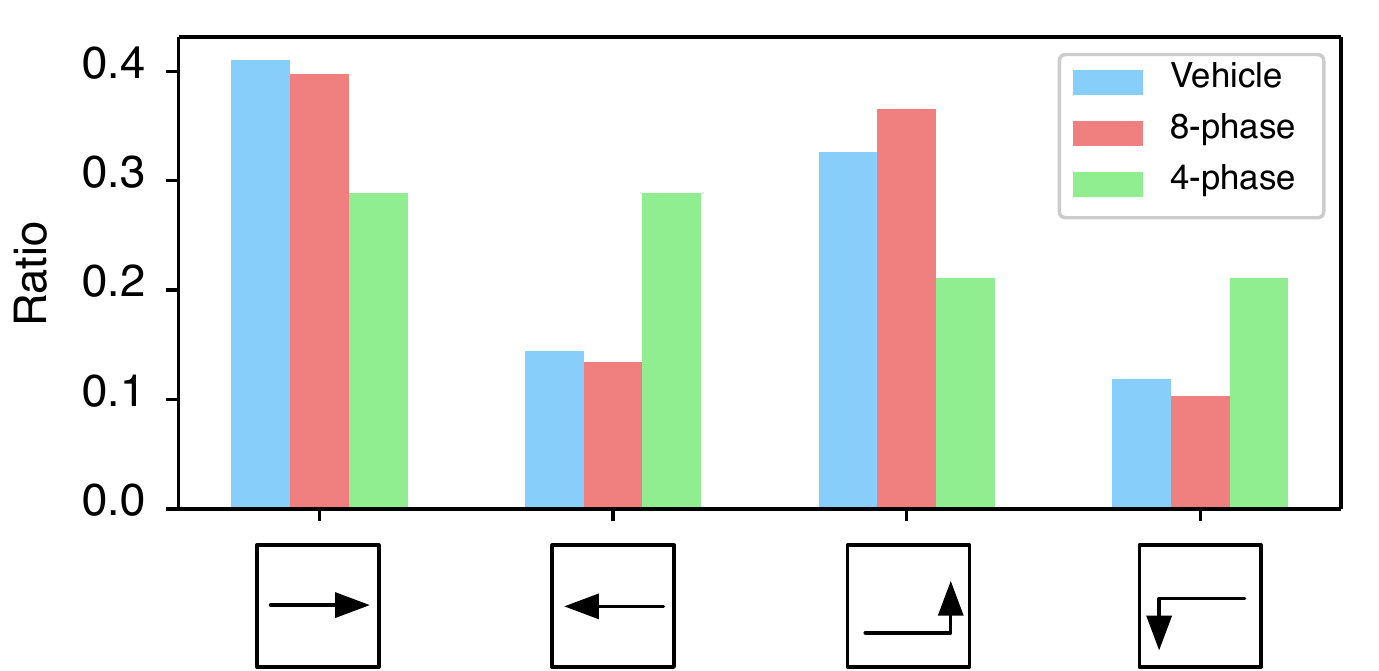}
\caption{Traffic volume ratio and green time ratio (under the 4-phase and 8-phase settings) of W and E approach movements at the intersection 1 in Jinan, China.}
\vspace{-1mm}
\label{fig:phase-setting}
\vspace{-3mm}
\end{figure}

As shown in Figure~\ref{fig:phase-setting}, when the vehicle volume is quite different on west and east approaches at the intersection 1 in Jinan, the policy learned under the 8-phase setting can adjust its green time accordingly, whereas the policy under 4-phase setting wastes a significant amount of time on the movement with light traffic(e.g., the east-through movement). Indeed, travel time is only $66.40$s under the 8-phase setting, but increases to $81.97$s under the 4-phase setting. Thus, the general 8-phase setting brings flexibility to traffic signal control as it adapts better to unbalanced traffic flows.

\subsection{Experiment on Different Environments}
\label{sec:complex-cases}

\subsubsection{Experiment on different intersection structures}
We also evaluate the performance of our method under different intersection structures (i.e., 3-, 4-, and 5-approach intersections).
For this experiment, we synthesize one typical traffic flow for each structure based on the mean and variance of traffic flows in Jinan, China. 

We modify our model as follows. For the 3-approach intersection, we disable some neurons in our network to make it compatible with this structure. In addition, the input is modified with zero padding for nonexistent movements.
For the 5-approach structure, we add another phase to the model, and then use the same process (i.e., phase demand modeling, phase pair embedding, and phase pair competition) as in Figure~\ref{fig:network} to predict the Q-value of all phases.

Detailed results of different intersection structures are reported in Table~\ref{tab:different-structures}. We can see that \ours performs consistently better than other methods and can be applied easily to all structures without major modification.

\begin{table}[tbhp]
\centering
\caption{Performance on different intersection structures.}
\label{tab:different-structures}
\begin{tabular}{c|c|c|c}
\toprule
Model & 3-approach & 4-approach & 5-approach \\
\midrule
\fixedtime & 166.16  & 93.21  & 211.26  \\
\formula & 159.12  & 67.17  & 231.33  \\
\SOTL & 123.43  & 65.39  & 124.71  \\
\hline
\NIPS & 108.94 & 125.84 & 140.33 \\
\deeplight & 108.27  & 60.38  & 151.92 \\
\hline
\ours & \textbf{81.57}  & \textbf{48.83}  & \textbf{110.66}  \\
\bottomrule
\end{tabular}
\label{table:overall}
\end{table}

\subsubsection{Extension to a multi-intersection environment.}
Compared with a single intersection, people sometimes concern more about the overall traffic light control for an area containing multiple intersections. A straightforward way to enable intelligent traffic light control in the  multi-intersection environment is to assign an independent RL agent for each intersection. 

\begin{figure}[htbp]
\centering
\vspace{-2mm}
\includegraphics[width=0.3\textwidth]{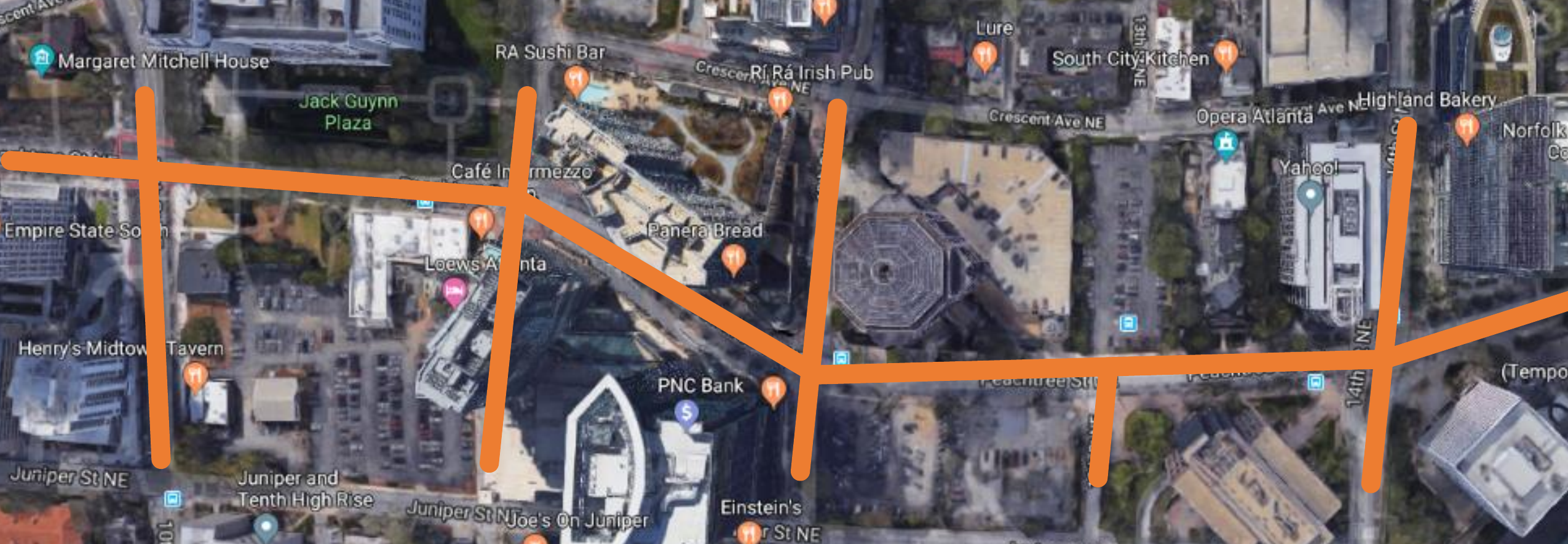}
\caption{Satellite image of a multi-intersection in Atlanta.}
\vspace{-2mm}
\label{fig:atlanta_map}
\end{figure}

To validate the potential of our model in the multi-intersection environment, we select a $3\times4$, a $4\times4$, and a $1\times5$ grid of intersections in Jinan, Hangzhou, and Atlanta respectively. The Jinan and Hangzhou data are selected from regions with relatively rich data coverage. Necessary missing data filling in are done to preprocess the data. The satellite image of the $1\times5$ grid in Atlanta is shown in Figure~\ref{fig:atlanta_map}. Performance of different methods is shown in Table~\ref{tab:multi-intersection}. We can see that \ours stands out among all traffic signal control methods again in this setting. For further improvement, coordination of neighboring intersections can be considered as a promising direction for future work.
\begin{table}[htbp]
\centering
\caption{Performance in a multi-intersection environment.}
\label{tab:multi-intersection}
\begin{tabular}{cccc}
\toprule
Model & Jinan & Hangzhou & Atlanta\\\midrule
\fixedtime & 880.18& 823.13 & 493.49\\
\formula & 385.46 & 629.77 &  831.34 \\
\SOTL & 1422.35 & 1315.98 & 721.15 \\
\hline
\NIPS & 1047.52 & 1683.05 & 769.46\\
\deeplight & 358.83 & 634.73 & 306.07 \\
\hline
\ours & \textbf{293.35} & \textbf{528.44} & \textbf{124.42} \\
\bottomrule
\end{tabular}
\end{table}

\subsection{Interpretation of Learned Policies}
To gain additional insight about what \ours has learned, we choose one intersection and visualize the learned policy in the following way: for each hour between 8 am and 8 pm, the busiest time in a day, we calculate the green light time assigned to each movement according to the specific policy and then normalize it to obtain the green time percentage. In the meantime, we compute the ratio of vehicle volume on each movement for reference. 

\begin{figure}[htbp]
\begin{center}
\includegraphics[width=0.36\textwidth]{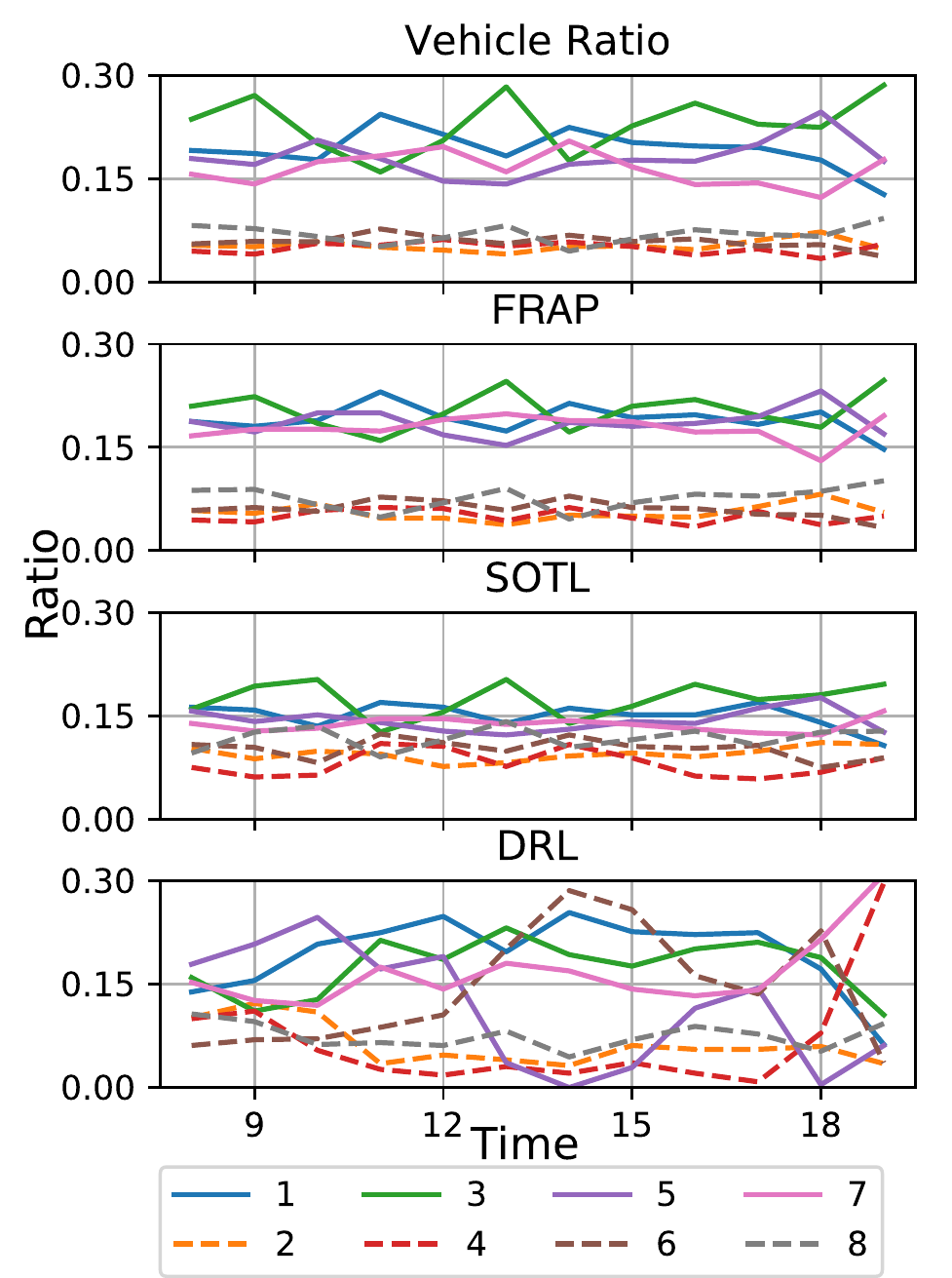}
\caption{Traffic volume ratio and green time ratio of three traffic signal control methods for each movement between 8 am and 8 pm. The number from 1 to 8 indicates a specific movement signal described in Figure~\ref{fig:phase}(b).}
\vspace{-3mm}
\label{fig:vehicle-ratio}
\end{center}
\vspace{-1mm}
\end{figure}

As shown in Figure~\ref{fig:vehicle-ratio}, the green time ratio of \ours synchronizes well with the percentage of traffic volume in each hour, while other baseline methods such as \SOTL and \NIPS would allocate green light time to each movement more randomly and irregularly. Specifically, during the selected period, the vehicle volume of four left-turn movements is relatively light compared with that of four through ones. Thus, an good policy is expected to assign more green time to the through movements. Figure~\ref{fig:vehicle-ratio} shows that \ours indeed divides the through and left-turn movements into two groups, whereas \SOTL and \NIPS largely mix them together.

%% file: conclusion.tex

\section{Conclusion}
\label{sec:conclusion}

In this paper, inspired by the universal principles of competition and invariance, we propose a novel RL model \ours for traffic signal control.
 We analyze the advantage of \ours over other RL methods in sampling efficiency and carry out comprehensive experiments on three datasets. Results demonstrate that our method converges faster and achieves better performance than state-of-the-art methods. Furthermore, we show the potential of our model in handling complex scenarios such as different intersection structures and multi-intersection environments. For future work, patterns of pedestrians and non-motorized vehicles need to be considered and a field study can be an important step for our model to get real-world feedback and for us to validate the proposed RL approach.

%% file: reference.tex

\bibliographystyle{ACM-Reference-Format}
  \bibliography{bibfile}